\documentclass[letterpaper, 10 pt, conference]{ieeeconf}  

\IEEEoverridecommandlockouts                              

\usepackage[noadjust]{cite}
\usepackage{graphicx}
\usepackage{booktabs}
\usepackage{multirow}
\usepackage{algorithm}
\usepackage{algorithmic}
\usepackage{hyperref}
\usepackage{pgfplots}
\usepackage{listings}
\usepackage{gensymb}
\usepackage{amsmath}
\pgfplotsset{compat=1.17}
\usepackage[table]{xcolor}

\title{\LARGE \bf
Open-Architecture End-to-End System for \\ Real-World Autonomous Robot Navigation
}

\author{
    Venkata Naren Devarakonda$^{*1}$, Ali Umut Kaypak$^{*1}$, Raktim Gautam Goswami$^{*1}$, Naman Patel$^{1}$,\\Rooholla Khorrambakht$^{1}$, Prashanth Krishnamurthy$^{1}$, Farshad Khorrami$^{1}$
    \thanks{*Equal Contribution}
    \thanks{$^{1}$Control/Robotics Research Laboratory (CRRL), Department of Electrical and Computer Engineering, NYU Tandon School of Engineering, Brooklyn, NY, 11201. E-mails: \{\texttt{d.naren, rgg9769, ak10531, nkp269, rk4342, prashanth.krishnamurthy,khorrami}\}@nyu.edu}
    \thanks{This work was supported in part by Google and in part by ARO under Grant W911NF-22-1-0028, and in part by the New York University Abu Dhabi (NYUAD) Center for Artificial Intelligence and Robotics (CAIR), funded by Tamkeen under the NYUAD Research Institute Award CG010.}
}

\begin{document}

\maketitle
\thispagestyle{empty}
\pagestyle{empty}

\begin{abstract}
Enabling robots to autonomously navigate unknown, complex, and dynamic real-world environments presents several challenges, including imperfect perception, partial observability, localization uncertainty, and safety constraints. Current approaches are typically limited to simulations, where such challenges are not present. 
In this work, we present a lightweight, open-architecture, end-to-end system for real-world robot autonomous navigation. Specifically, we deploy a real-time navigation system on a quadruped robot by integrating multiple onboard components that communicate via ROS2. Given navigation tasks specified in natural language, the system fuses onboard sensory data for localization and mapping with open-vocabulary semantics to build hierarchical scene graphs from a continuously updated semantic object map. An LLM-based planner leverages these graphs to generate and adapt multi-step plans in real time as the scene evolves.
Through experiments across multiple indoor environments using a Unitree Go2 quadruped, we demonstrate zero-shot real-world autonomous navigation, achieving over 88\% task success, and provide analysis of system behavior during deployment.
\end{abstract}

\section{Introduction}
\begin{figure*}[!ht]
    \centering
    \includegraphics[width=0.95\linewidth]{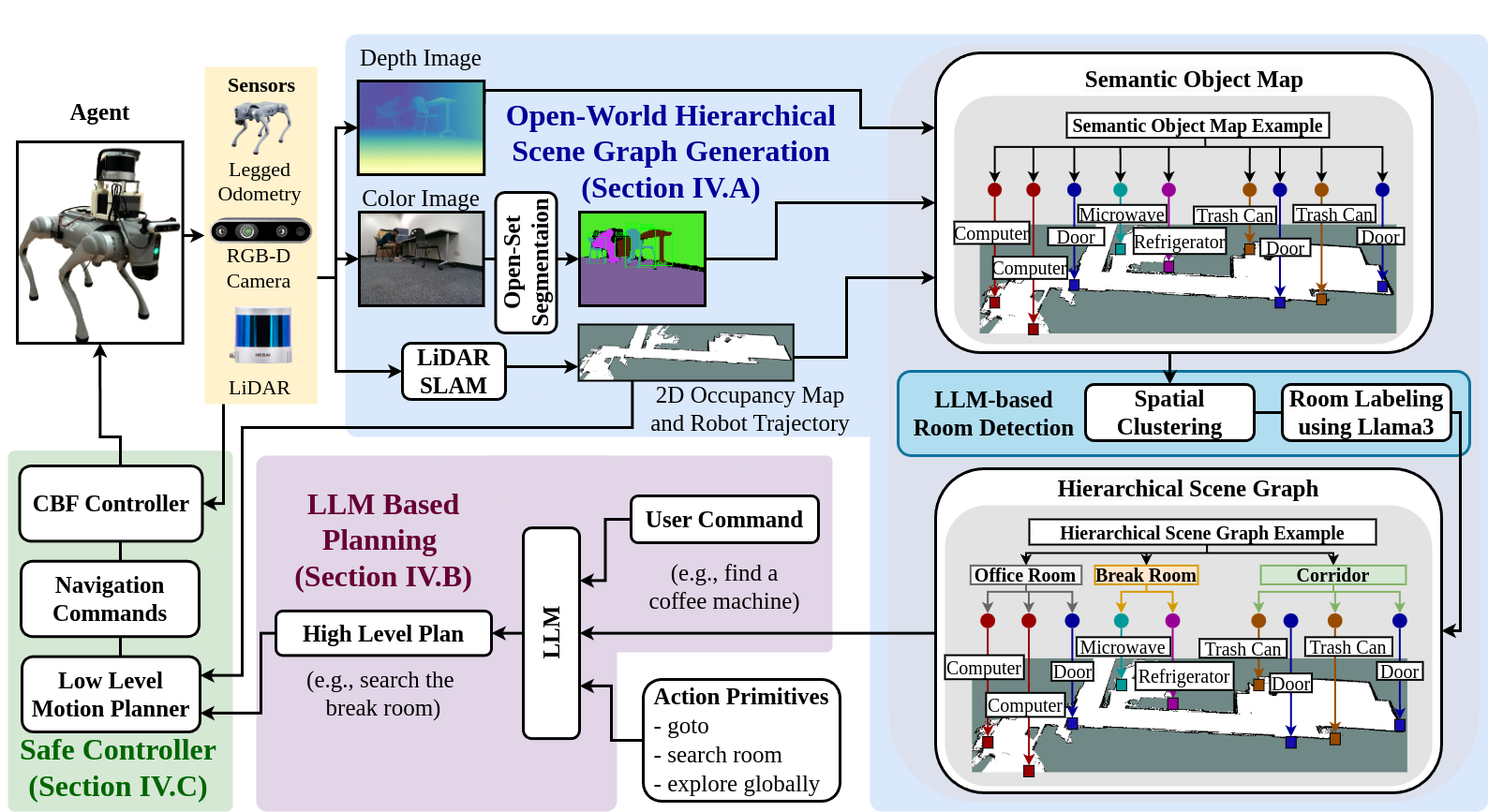}
    \caption{We introduce \textbf{OrionNav}, a modular open-architecture framework for end-to-end real-world autonomous navigation, deployed on a Unitree Go2 quadruped robot. OrionNav fuses data from onboard LiDAR and odometry sensors for robust localization and mapping, while integrating open-world semantics to produce a \textbf{semantic object map} of the environment. This map is then clustered into distinct rooms, and room labels are assigned using the LLaMA3 LLM, generating a \textbf{hierarchical scene graph}. An \textbf{LLM-based planner} utilizes this scene graph, along with user commands, to create a high-level task execution plan, guiding low-level controllers to safely and efficiently achieve designated goals.}
    \label{fig:flow}
    \vspace*{-0.5cm}
\end{figure*}

As agentic artificial intelligence systems advance, enabling their practical deployment in real-world tasks like navigation, manipulation, and interaction becomes increasingly important~\cite{ZhangHXZ22,duan2022survey}. While prior work has made progress in this direction~\cite{yokoyama2024vlfm,Maggio2024Clio,liu2024okrobot,gu2024conceptgraphs}, it remains limited, often confined to simulation, characterized by long execution times, storage-intensive representations, or an inability to handle dynamic changes, and reliant on simplifying assumptions such as perfect perception and localization.
These limitations hinder real-world deployment, particularly for robots that must operate in real time with onboard compute and generalize zero-shot to new environments. In practice, such deployment requires (a) reliable onboard mapping and localization, (b) an intelligent planner capable of online adaptation, and (c) a safe low-level controller for execution.
Some recent methods (e.g.,~\cite{yokoyama2024vlfm, rajvanshi2024saynav}) demonstrate real-world examples, but they are not end-to-end in the real-world settings and rely on proprietary modules for tasks such as localization and navigation.
In this work, we present \textit{OrionNav}, the first open-architecture, end-to-end system for real-time autonomous navigation deployed on a real-world robot. The system integrates open-vocabulary hierarchical scene graphs for mapping, an LLM-based planner for high-level reasoning, and a control barrier function (CBF)-based low-level controller for safe execution (Fig.~\ref{fig:flow}).

Mapping and localization are fundamental capabilities for autonomous robots, commonly addressed through Simultaneous Localization and Mapping (SLAM) methods~\cite{WangL24f,YinXLCXSSW24}. Recent work~\cite{RosinolVAHCSGC21,WangTCXD24,goswami2024floor,WuLXYDY0ZT0GT24} augments SLAM with object detection and segmentation, including open-vocabulary perception, to construct rich semantic maps. Building on these, we construct hierarchical scene graphs that encode objects and their relationships, providing structured scene representations for planning. To enable real-time onboard deployment, we adopt a lightweight design that combines 2D LiDAR SLAM~\cite{slam_toolbox} with an optimized open-vocabulary segmentation model running on NVIDIA Jetson compute. Detected objects are represented as points within the scene graph, enabling efficient construction and updates during operation. These representations provide the structured context required for high-level planning with LLMs.

LLMs have recently emerged as powerful planners due to their ability to reason over structured, context-rich representations~\cite{10161317,rajvanshi2024saynav,llm_planning_survey}. In our system, the LLM uses hierarchical scene graphs, formatted as JSONs, to generate high-level plans for navigation tasks specified in natural language. Because the scene representation is continuously updated, the planner can revise plans online as the environment evolves. These high-level plans are executed through lower-level planners and controllers responsible for motion generation and CBF-based obstacle avoidance. The components communicate through ROS2, enabling the robot to translate natural language instructions into safe physical actions and navigate to target objects or locations in previously unseen environments. All modules except the LLM planner run on NVIDIA Jetson computers onboard the robot.

In summary, we present the first open-architecture, end-to-end system for real-world autonomous robot navigation. The system comprises an open-vocabulary hierarchical scene graph for semantic mapping, an LLM-based planner for adaptive high-level decision-making, and a safety-critical low-level controller for execution. It is fully deployed on a Unitree Go2 quadruped robot, operating in real time using onboard sensing and computation.


\section{Related Works}
\label{sec:related-works}
\subsection{Scene Graphs and Semantic Mapping}
\label{sub sec: scene-graph}
Scene graphs provide an efficient representation for robotics by encoding environments as nodes (objects/regions) connected by edges (spatial/semantic relationships). Recent work integrates open-vocabulary semantics using CLIP~\cite{RadfordKHRGASAM21} embeddings and VLMs~\cite{achiam2023gpt}, enabling scene understanding beyond predefined categories. While many scene graph approaches exist, the most related to our work include  ConceptGraphs~\cite{gu2024conceptgraphs}, HOV-SG~\cite{werby2024hierarchical}, and Clio~\cite{Maggio2024Clio}, which 
create detailed scene representations for planning and navigation. However, these methods face limitations for real-time deployment: ConceptGraphs relies on offline processing, HOV-SG uses computationally expensive clustering, and Clio is restricted to predefined task sets.

Semantic scene graph integrates semantic segmentation with SLAM, providing both geometric structure and semantic understanding for robot navigation~\cite{duan2022survey}. Recent advances in vision–language models have enabled open-vocabulary segmentation through large-scale image–text pretraining~\cite{WuLXYDY0ZT0GT24}. Methods such as Open-Vocabulary SAM~\cite{YuanLZLC24} improve segmentation accuracy and efficiency, while FC-CLIP~\cite{fcclip} introduces a fully convolutional architecture that reduces computational complexity while maintaining strong zero-shot capabilities.
In 3D perception, systems such as Kimera~\cite{RosinolVAHCSGC21} construct dynamic scene graphs by combining semantic segmentation with SLAM. Other approaches process point clouds directly for open-vocabulary 3D understanding~\cite{DingYXZBQ23,SchultEHLTL23}. However, these methods struggle with computational efficiency in larger environments, often requiring high-end GPUs unsuitable for mobile robots.

\subsection{VLM/LLM for Robotic Task Planning and Navigation}
\label{sub sec: task-planning}
Recent advances in robotic task planning leverage LLMs for navigation and manipulation tasks~\cite{10161317, huang2022inner, Ahn2022DoAI, 11128486}. 
ConceptGraphs \cite{gu2024conceptgraphs}, while focusing primarily on scene graph generation, employs an LLM-based planner to retrieve target objects from a pre-built scene graph as a downstream task. Similarly, HOV-SG \cite{werby2024hierarchical} emphasizes scene graph generation and utilizes an LLM to parse complex natural language queries for object retrieval from the pre-built scene graph. 
While SayPlan \cite{rana2023sayplan}, MoMa-LLM \cite{10632580}, and SayNav \cite{rajvanshi2024saynav} also leverage LLMs as high-level planners, they face constraints that limit their practical application. SayPlan \cite{rana2023sayplan} requires pre-built scene graphs, whereas MoMa-LLM \cite{10632580} and SayNav \cite{rajvanshi2024saynav} rely on known robot poses and semantic object masks. OK-Robot~\cite{liu2024okrobot} uses CLIP for open-ended pick-and-place tasks, while PixelNav~\cite{cai2024bridging} enables RGB-based waypoint generation. However, these methods either create non-reusable environment representations or require pre-built scenes. Recent VLM-enhanced navigation includes VLFM~\cite{yokoyama2024vlfm} for semantic value maps, ESC~\cite{zhou2023esc} leveraging commonsense knowledge, and StructNav~\cite{Chen-RSS-23} combining geometric frontiers with semantic filtering. GAMAP~\cite{huang2024gamap} introduces multi-scale geometric-affordance guidance, while Zero-shot navigation~\cite{wen2025zero} employs two-step VLM reasoning for exploration. However, dense semantic representations in such methods limit map reusability and require high-end GPUs for onboard inference.
In contrast, OrionNav overcomes these constraints by performing perception, planning, and control for navigation tasks using onboard sensors in real-time, enabling autonomous navigation in diverse, unknown, and dynamic real-world environments.

\section{Framework}
\label{sec:framework}

\noindent\textbf{Problem Statement.}
We consider the problem of deployment of embodied agents in the real world for goal navigation. Specifically, we want to enable a mobile robot equipped with standard onboard sensors like cameras and LiDARs to interpret a high-level object/room search task and autonomously and safely navigate the environment to reach the target.  Real-world deployment introduces several challenges that are often abstracted away in simulation, including imperfect perception, partial observability, localization uncertainty, and safety constraints during navigation. Addressing these challenges requires a tightly integrated system capable of mapping the environment, understanding its semantic structure, and generating safe navigation commands based on high-level task objectives.


\vspace{1.2mm}
\noindent\textbf{Framework Overview.}
We introduce OrionNav, a modular open-architecture framework that integrates the algorithmic components required to deploy and evaluate object navigation methods on physical robots in real-world environments. The framework provides simple, explainable baseline modules that together form a robust navigation pipeline while remaining flexible enough to support alternative algorithms. The system processes real-time sensor data and runs multiple modules in parallel to search for and navigate to a target object or room.

A LiDAR-based 2D SLAM module in OrionNav localizes the robot and builds a map of the explored environment, providing global spatial context. A semantic perception module detects objects from RGB-D observations and grounds them in the map. These objects are organized into a hierarchical scene graph that associates them with the rooms they belong to, enabling higher-level reasoning about the environment. The graph is provided to an LLM planner to generate high-level commands, which are executed through a navigation stack combining goal-directed navigation, frontier-based exploration, and a safe control wrapper for reliable motion in complex environments.
All components communicate through ROS2 in real time, enabling end-to-end execution on robotic platforms. While the framework provides a complete pipeline, it remains highly modular, allowing individual modules to be replaced with alternatives. This design makes OrionNav a practical platform for evaluating and comparing object navigation approaches in real-world settings.

\subsection{Open-World Hierarchical Scene Graph}
\label{sec:scenegraph}
\noindent \textbf{Open-Vocabulary Semantic Segmentation.}
We obtain semantic object masks from RGB frames using an open-vocabulary segmentation model based on FC-CLIP~\cite{fcclip}. The model uses a frozen ConvNeXt-L~\cite{convnext} CLIP backbone to extract image features and predict candidate object masks, which are then classified using text embeddings of object categories. These embeddings are generated from template-based prompts for each class and its synonyms, enabling zero-shot recognition of unseen objects~\cite{fcclip}. 
To improve robustness, the system combines scores from an in-vocabulary classifier for categories seen during FC-CLIP training and an out-of-vocabulary classifier that preserves CLIP’s recognition for unseen classes, filtering predicted masks by confidence before semantic object mapping.
Synonyms are mapped to a canonical class label to maintain consistent object naming for downstream scene graph construction.

To enable real-time onboard inference on our resource-constrained quadruped, we introduce several optimizations to the FC-CLIP pipeline. The model is converted to half-precision (FP16) to reduce memory usage, and custom CUDA kernels are implemented for the multi-scale deformable attention module to address numerical instability in reduced precision. We also replace the standard softmax with a more efficient approximation and vectorize mask prediction by flattening tensors for parallel top-k selection. These modifications significantly improve runtime efficiency while maintaining segmentation performance, enabling real-time operation using the robot's onboard compute.

\begin{figure}[t]
    \includegraphics[width=\linewidth]{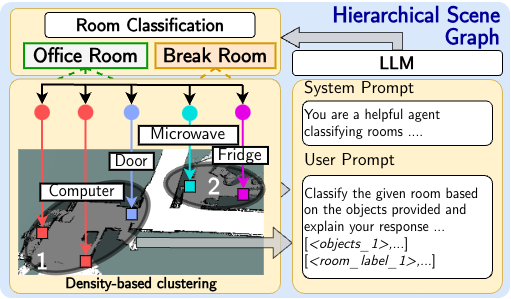}
    \caption{The semantic construct of indoor spaces enable the abstraction of rooms from the detected objects in the scene. The objects are first clustered based on their concentration in the map. For each cluster, an LLM queried to predict the room label given the object classes within and a set of feasible candidate room labels. The generated hierarchical graph serves as grounded semantic context for the LLM-based planner.}
    \label{fig:scene_graph}
    \vspace*{-0.3cm}
\end{figure}

\vspace{1.2mm}
\noindent \textbf{Semantic Object Mapping.}
We use the off-the-shelf SLAM toolbox~\cite{slam_toolbox} to estimate the robot’s pose and construct a 2D occupancy grid map of the environment. The module leverages onboard LiDAR and legged-inertial odometry from the Unitree Go2 and applies pose-graph optimization to maintain consistent 2D localization and mapping during operation.
Detected objects are grounded in the map by reprojecting segmentation mask centers into 3D using the depth frame and camera intrinsics, then transforming them to the world frame using the SLAM-estimated robot pose. Object centers are associated with existing instances using density-based clustering (DBSCAN), enabling consistent tracking across frames and removing duplicates. If no match is found, a new object instance is created. When pose-graph updates occur (e.g., from loop closure), object locations are updated to remain aligned with the refined map.

\vspace{1.2mm}
\noindent \textbf{Hierarchical Scene Graph.}
\label{sec:hierarchical_scene_graph}
Indoor environments are inherently structured into distinct semantic components, primarily floors and rooms such as `office,' `kitchen,' and `bedroom,' based on their intended function. The consistency of these labels across indoor environments makes them highly informative semantic descriptors. This semantic structure naturally lends itself to representation through hierarchical scene graphs, as explored in prior work~\cite{Maggio2024Clio,werby2024hierarchical}.

Leveraging this structure, we partition the scene into multiple spatial areas by applying DBSCAN on the coordinates of mapped objects. For each resulting cluster, we infer a semantic room label using a large language model (LLM). The LLM receives the list of object labels within the cluster along with a curated set of candidate room labels and guidelines to ensure consistent labeling across the environment. Room labels are generated in real time on-device using a memory-efficient 4-bit quantized version of LLaMA 3~\cite{llama3}. The resulting representation is stored as a hierarchical scene graph implemented as a tree structure, where each node contains properties including its hierarchy level, semantic label, class-specific instance identifier relative to the parent node, spatial location, and parent-child relationships (Fig.~\ref{fig:scene_graph}). This representation provides an interpretable global semantic map that enables the language-based high-level planner to generate effective search strategies.

\subsection{LLM Planner}
\label{sec:llm_planner}
The high-level planning module uses GPT-4-Turbo~\cite{achiam2023gpt} to generate navigation and interaction strategies for the robot. The planner operates using three action primitives: (a) \textit{goto}, which directs the robot to a specified room or object; (b) \textit{search\_room}, which instructs the robot to navigate to a room and scan for additional objects; and (c) \textit{explore\_globally}, which initiates autonomous exploration when insufficient information about the environment is available.

The LLM receives the task description together with the hierarchical scene graph (Sec.~\ref{sec:hierarchical_scene_graph}), formatted as JSON, which contains the discovered rooms and objects in the environment. The prompt specifies the LLM’s role, available action primitives, and guidelines for reasoning and output formatting. Based on this information, the LLM selects an appropriate action primitive and fills in the required arguments corresponding to rooms or objects in the scene graph. The planner also generates feedback explaining the selected action.
Planning proceeds iteratively, as shown in Algorithm~\ref{alg:task_execution}. If the scene graph is empty, the system first performs a 360$^\circ$ scan to initialize observations. The LLM then generates a high-level action conditioned on the current scene graph and task. After action executing, the scene graph is updated using new observations, and the LLM produces the next action until the task is completed. Each action is executed through the safe robot controller (Sec.~\ref{sec:navigation}).

\begin{algorithm}
\caption{Task Execution (\textbf{Input:} user\_command)}
\label{alg:task_execution}
\begin{algorithmic}[1] 
    \STATE action\_history $\gets$ []
    \STATE feedback $\gets$ None; task\_accomplished $\gets$ False
    \WHILE{not task\_accomplished}
    \STATE scene\_graph $\gets$ \textit{HierarchicalSceneGraph.get\_latest()} 
    \STATE action $\gets$ \textit{LLMPlanner.call(user\_command, \\scene\_graph, feedback, action\_history)}
    \STATE action\_history.append(action)
    \STATE action\_success $\gets$ \textit{RobotApi.send(action)}
    \IF{action\_success}
        \IF{action = goto(\textless arg\textgreater)}
         \STATE task\_accomplished $\gets$ \textit{ask\_agent(`Task done?')}
         \IF{not task\_accomplished}
         \STATE feedback $\gets$ `Task has not been accomplished.' 
        \ELSE
         \STATE feedback $\gets$ None
        \ENDIF
        \ELSE
        \STATE feedback $\gets$ `Task has not been accomplished.' 
        \ENDIF
        \ELSE
        \STATE feedback $\gets$ \textit{RobotApi.get\_recent\_action\_logs()}
    \ENDIF
    \ENDWHILE
\end{algorithmic}
\end{algorithm}

\subsection{Safe Robot Controller}
\label{sec:navigation}
\noindent \textbf{Navigation.} The ROS2 navigation stack~\cite{nav2} executes the \textit{goto} commands generated by the LLM planner. Global trajectory planning uses a cost-aware $A^*$ search over an occupancy grid map built from a SLAM map (Sec.~\ref{sec:hierarchical_scene_graph}). To improve safety, obstacles in the grid are radially inflated to account for the robot’s footprint and maintain a safety margin. For local execution, we use a sampling-based Model Predictive Path Integral (MPPI) controller that accounts for the robot’s nonlinear dynamics while maintaining a forward-looking planning horizon. The MPPI controller generates velocity commands passed to the robot’s locomotion controller.

\vspace{1.2mm}
\noindent \textbf{Control Barrier Function.} 
While the MPPI controller accounts for nearby obstacles through the cost map, it does not inherently guarantee collision avoidance. To further enhance safety, we incorporate a Control Barrier Function Quadratic Programming safety filter~\cite{dai2024sailing} between the MPPI output and the robot’s low-level velocity tracker. The filter processes raw LiDAR point cloud data to assess the immediate surroundings and adjusts control commands, when necessary, to proactively avoid collisions. This additional safety layer improves robustness in complex and dynamic environments.

\vspace{1.2mm}
\noindent \textbf{Exploration.} To facilitate efficient exploration of unknown environments, we utilize the \textit{m-explore} ROS2 package\footnote{https://github.com/robo-friends/m-explore-ros2}. Each frontier is weighted based on its distance $\mathcal{F}_d$ from the robot and its size $\mathcal{F}_n$, which reflects the potential unexplored area reachable from that frontier. The frontier prioritization weight is defined as $\mathcal{F}_w = \alpha\mathcal{F}_d + \mathcal{F}_n$. We set $\alpha=50$ in our experiments to favor closer frontiers and improve robustness to measurement errors in long-range retrieval tasks.

\section{Experiments}
\label{sec:experiments}

\subsection{Implementation Details}
\label{sec:implemetation}

The algorithm and all its components are implemented on the Unitree Go2 quadruped robot shown in Fig.~\ref{fig:flow}. In addition to its built-in legged-inertial sensors, the robot is equipped with a top-mounted Hesai LiDAR for high-resolution mapping and a RealSense D455 stereo camera. Except for the LLM planner’s cloud API calls, all processes run on two onboard NVIDIA Jetson devices: the AGX Orin handles computationally intensive semantic tasks, while the Orin Nano manages high-frequency SLAM, navigation, and control.
During operation, RGB input from the stereo camera (30 Hz, 640$\times$480) is processed by the open-vocabulary segmentation model to generate object masks at about 10 Hz. Our model recognizes 560 object classes and processes 1306 text categories, while allowing new classes to be added dynamically. The semantic object map is updated at 2 Hz by integrating odometry, stereo depth, segmentation outputs, and pose-graph corrections, while the hierarchical scene graph is refreshed at rates up to 0.5 Hz depending on the number of detected objects and rooms.


\subsection{Experiment Setup}
To demonstrate the framework’s zero-shot object navigation capabilities in real-world settings, we evaluate it in four different office environments (Sec.~\ref{sec:results}). We further conduct comparative experiments with two related methods and ablated versions of OrionNav (Sec.~\ref{sec:comparisons}) to highlight how our design choices enable reliable real-world deployment. These experiments also highlight the value of an open-sourced, end-to-end modular architecture towards efficient testing and comparison of different methods in real-world settings.

The office experiments are organized into three scenarios. Scenarios A and B focus on object retrieval tasks, while Scenario C addresses room navigation. In Scenario A (short-range object retrieval), target objects are placed in the same room as the robot’s starting position, testing the system’s ability to quickly identify and approach objects with minimal navigation complexity. Scenario B (long-range object retrieval) presents a more challenging setup, requiring navigation through corridors and multiple rooms to locate the target. Scenario C (room navigation) evaluates the robot’s ability to interpret higher-level goals by detecting and navigating to a specified room.
 

\begin{table}[t]
\centering
\caption{Experiment Scenarios and Results}
\label{tab:exp_setup}
\begin{tabular}{@{}ccc@{}}
\toprule
\textbf{Task Type}                                                      & \textbf{Example Prompts}                                                                                         & \textbf{\begin{tabular}[c]{@{}c@{}} No. of Exp \\ (Success/Total) \end{tabular}} \\ \midrule
\begin{tabular}[c]{@{}c@{}}\textbf{Scenario A}: Short \\ Range Object Retrieval\end{tabular} & \multirow{2}{*}{\begin{tabular}[c]{@{}c@{}}find a potted plant, \\ go to a monitor,\\ find a chair\end{tabular}} & 32/36                   \\ \cmidrule(r){1-1} \cmidrule(l){3-3} 
\begin{tabular}[c]{@{}c@{}}\textbf{Scenario B}: Long \\ Range Object Retrieval\end{tabular}  &                                                                                                                  & 41/45                   \\ \midrule
\textbf{Scenario C}: Room Navigation                                                         & find a break room                                                                                                & 12/15                   \\ \midrule
\textbf{Total }                                                                  &                                                                                                                  & \textbf{85/96}                   \\ \bottomrule
\end{tabular}
\end{table}

\begin{figure}[!th]
    \centering
    \includegraphics[width=1\linewidth]{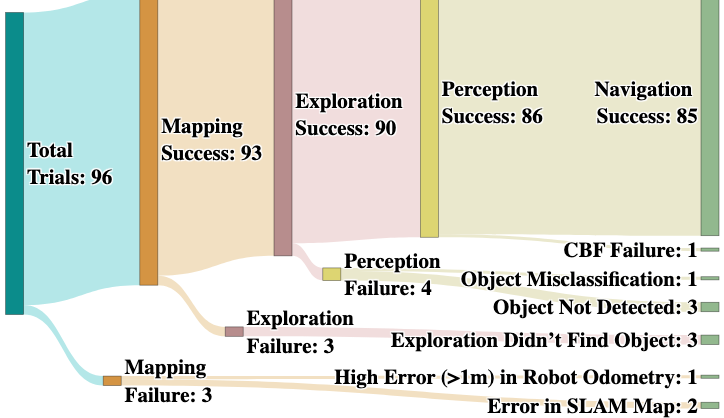}
    \caption{\textbf{Success and failure cases of OrionNav across all experiments}. Visualization of our task execution framework with breakdown of different observed failures that occur due to perception, and navigation failures. A detailed breakdown of failures provided on the right.}
    \label{fig:results}
\end{figure}

\subsection{Results}
\label{sec:results}
OrionNav demonstrates strong real-world applicability, successfully completing 85 of 96 trials (Table~\ref{tab:exp_setup}). Fig.~\ref{fig:results} summarizes the results, including failure cases and their causes. Two failures were due to erroneous SLAM maps and one due to high odometry error; excluding these, the remaining 93 trials proceeded without localization issues.
Among them, the exploration algorithm failed to guide the robot to the target location in three trials. Of the 90 trials with successful exploration, four failed due to semantic segmentation errors where the target object was either undetected or misclassified. Additionally, one failure occurred in the CBF-based navigation system. Notably, no failures were attributed to the LLM planner or the scene graph generator, highlighting their robustness. Next, we provide a detailed analysis of the performance across the three task scenarios.


\vspace{1.2mm}
\noindent \textbf{Scenario A.}
The robot and target object are placed in the same region without occlusions. We conduct 36 experiments using four distinct objects placed at different locations. To account for the stochasticity of LLM outputs, three trials are performed for each target location from the same initial position. Our framework achieves a high success rate of approximately 88\% in this scenario. During the robot’s initial 360$^\circ$ scan, the system detects most objects in the environment and assigns room labels. If the target is observed during this scan, the planner immediately issues a \textit{goto} command to navigate to it; otherwise, it issues a \textit{search\_room} command. This adaptive strategy proves effective, as the target object is typically detected during the subsequent search phase, demonstrating the robustness of the exploration approach.

\begin{figure*}[!ht]
    \centering
    \includegraphics[width=0.97\linewidth]{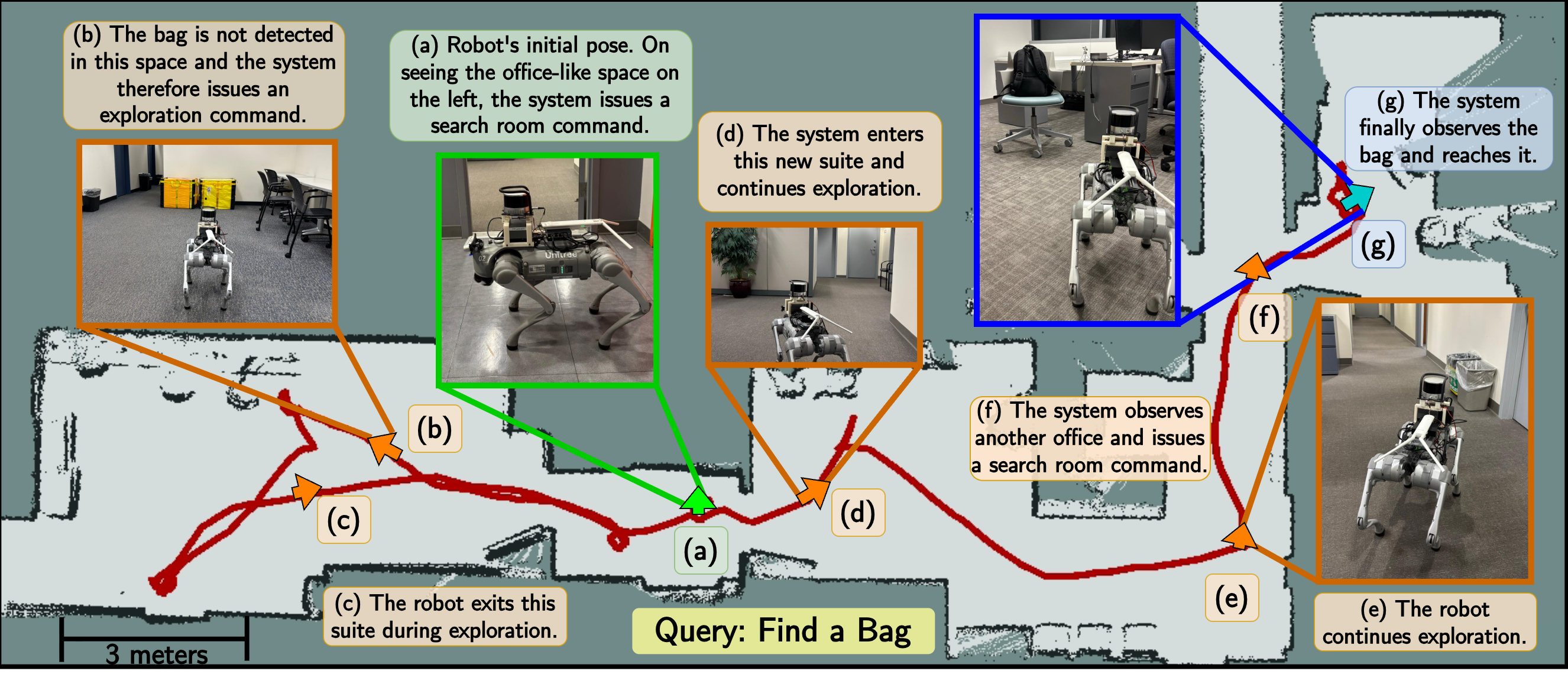}
    \caption{\textbf{Long range object navigation task}: The system is tasked with locating a bag on a floor of a building. It begins at the starting position (a) and initially searches the office-like space (b). After failing to find the bag, the system initiates global exploration (c, d, e). Upon detecting another office space (f), the system enters and searches it (g), successfully locating the bag.}
    \label{fig:scenario-B2}
\end{figure*}

\vspace{1.2mm}
\noindent \textbf{Scenario B.}
In this scenario, the target object is placed farther from the robot, often in separate rooms or corridors. Without prior knowledge of the environment, the robot must explore in real time, making the task more challenging than Scenario A. We use different objects placed at 15 distinct target locations, performing three trials from similar starting positions for each location.
Despite the increased complexity, the framework achieves a success rate of over 91\%. An example of Scenario B execution is shown in Fig.~\ref{fig:scenario-B2}, where the robot performs global exploration multiple times until it detects an office, enters it, and searches the room to successfully locate the target bag.

\begin{figure*}[!ht]
    \centering
    \includegraphics[width=\linewidth]{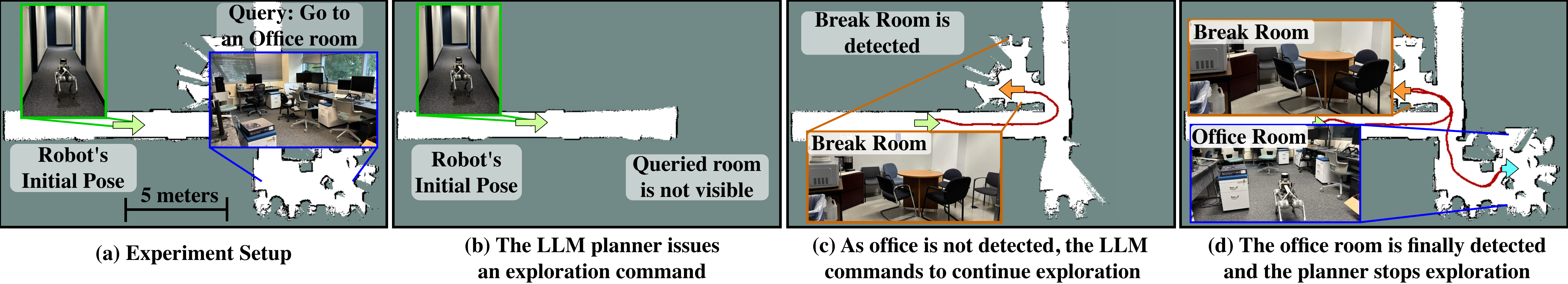}
    \caption{\textbf{Room navigation task:} The system is tasked with locating an office room on a building floor. The LLM planner initiates the task by issuing an exploration command, during which the robot first encounters a break room. Continuing its search, the robot eventually reaches an office room which is correctly detected and classified during scene graph generation. The LLM planner then halts exploration, successfully completing the task.}
    \label{fig:scenario-C}
    \vspace*{-0.2cm}
\end{figure*}


\vspace{1.2mm}
\noindent \textbf{Scenario C.}
Scenario C focuses on locating specific rooms. We evaluate four target rooms from different starting positions, repeating each experiment three times with slight variations, resulting in 15 trials with 12 successes. Fig.~\ref{fig:scenario-C} shows an example where the system searches for an office not visible from the robot’s starting position. The LLM planner first issues an exploration command (Fig.~\ref{fig:scenario-C}b). During exploration, the robot enters a room with a coffee maker, microwave, refrigerator, table, and chairs, which is classified as a break room (Fig.~\ref{fig:scenario-C}c). Since the target is not found, exploration continues. The robot then enters another room containing computers, monitors, tables, cabinets, and chairs (Fig.~\ref{fig:scenario-C}d). Once classifying this as an office, the system halts exploration and completes the task.

\subsection{Comparative Studies}
\label{sec:comparisons}

While OrionNav's contribution lies in its open-architecture and end-to-end real world deployment, this section provides qualitative comparisons with two related methods and two ablated versions of OrionNav to highlight how our design choices across modules enable this deployment.

A closely related approach to OrionNav is VLFM~\cite{yokoyama2024vlfm}, an object navigation method that uses vision-language models for exploration and is primarily evaluated in simulation. We use VLFM as a representative benchmark for zero-shot camera-based object navigation methods developed in simulated settings. In our real-world tests, VLFM was unable to complete the tasks considered, highlighting challenges often encountered when transferring simulation-focused approaches to physical robots. These include reliance on accurate odometry without SLAM-based correction, predefined map sizes, limited field-of-view mapping (front-view only) combined with frontier exploration, and pre-trained navigation policies that may not generalize well to new environments. Increasing map size can partially address coverage limitations but introduces computational trade-offs. For example, matching the size of our test environment required increasing VLFM’s map size to 2.5$\times$ the default, reducing mapping frequency by 37.5\% (from $\sim$0.9 Hz to 0.5 Hz). Increasing it to 5$\times$, roughly the size of a building floor, reduces the frequency to 0.17 Hz due to the quadratic scaling of dense map computation. Consistent with these challenges, when we emulate realistic localization noise in the HM3D simulation suite by injecting zero-mean instantaneous and accumulated Gaussian noise into position and yaw (position: $\sigma_{\text{inst}}=0.015$ m, $\sigma_{\text{acc}}=0.0075$ m; yaw: $\sigma_{\text{inst}}=0.015$ rad, $\sigma_{\text{acc}}=0.0075$ rad), VLFM’s success rate drops by 10.2\% (from 52.5\% to 42.3\%). In contrast, OrionNav integrates SLAM-based localization, wide-field semantic perception, and modular navigation components designed for onboard operation, enabling more reliable deployment on real robots.

We further compare our hierarchical scene graph construction with Clio~\cite{Maggio2024Clio} by integrating it into our system. In the original work, Clio was evaluated on an RTX 4090 GPU, which provides roughly 16$\times$ more FLOPs than the NVIDIA Orin onboard our robot. Due to its computational demands, Clio could not run in real time on the robot. When evaluated offline on recorded sensor data, it produced frequent object misclassifications and false positives. It also tended to fragment single rooms into multiple segments, leading to inconsistent room detection and unsuccessful navigation to the target object. OrionNav, on the other hand, uses a lightweight hierarchical scene graph construction method designed for real-time onboard operation, producing more consistent room-object associations and enabling reliable task execution on the robot.

Finally, we evaluate OrionNav against two ablated versions of itself under dynamic environmental changes. In each experiment, the robot searches for an object that is initially absent and introduced later. We test two settings across nine trials: one using a pre-built map and another where the robot builds the map from scratch. In both cases, the queried object is added after the mapping phase, and the goal is to locate it while minimizing task completion time. We compare our framework with two baselines: \textit{Frontier-Search} and \textit{Object-Map-Search}. \textit{Frontier-Search} explores frontiers and switches to random search once all frontiers are exhausted. Since it lacks semantic reasoning, an operator manually checks the object map to determine whether the queried object has been detected. \textit{Object-Map-Search} uses a semantic object map without room labels and introduces a new action primitive, \textit{search\_object(\textless object name\textgreater)}, which navigates to known instances of the object and performs a 360$^\circ$ scan to detect nearby objects. Table~\ref{tab:comparison_experiment_duration} reports task completion times, with tasks exceeding 10 minutes considered failures.

\begin{table}[t]
\centering
\caption{\textbf{Comparisons with Baseline Methods:} Task completion times (in minutes) for each method across experiments (E1-E9). Methods that exceed 10 minutes are considered failures (F).}
\vspace*{-0.3cm}
\label{tab:comparison_experiment_duration}
\setlength{\tabcolsep}{4pt}
\begin{tabular}{lccccccccc}
\toprule
\textbf{Method} & \textbf{E1} & \textbf{E2} & \textbf{E3} & \textbf{E4} & \textbf{E5} & \textbf{E6} & \textbf{E7} & \textbf{E8} & \textbf{E9} \\
\midrule
Frontier-Search   & \cellcolor{gray!20}\textbf{3} & F & F & F & F & F & F & F & F \\
Object-Map-Search & F & F & F & 3 & \cellcolor{gray!20}\textbf{2} & 3 & 4 & F & 5 \\
\textbf{OrionNav (Ours)}    & 5 & \cellcolor{gray!20}\textbf{3} & \cellcolor{gray!20}\textbf{4} & \cellcolor{gray!20}\textbf{2} & \cellcolor{gray!20}\textbf{2} & \cellcolor{gray!20}\textbf{2} & \cellcolor{gray!20}\textbf{3} & \cellcolor{gray!20}\textbf{3} & \cellcolor{gray!20}\textbf{4} \\
\bottomrule
\end{tabular}
\vspace*{-0.45cm}
\end{table}

\begin{figure}[!ht]
    \centering
    \includegraphics[width=0.88\linewidth, trim={0 0 0 0}, clip]{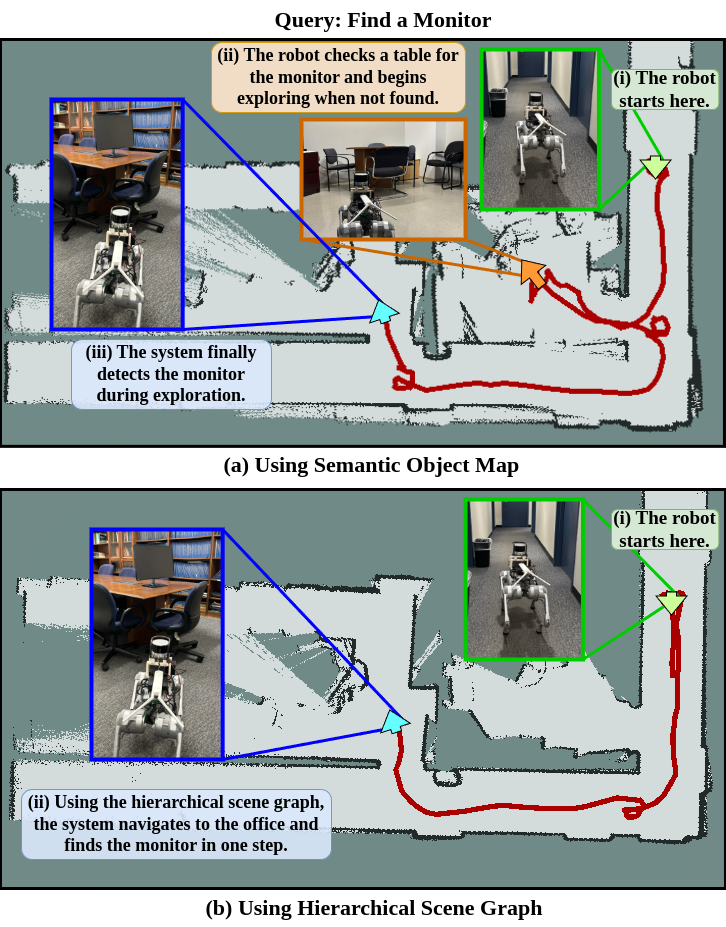}
    \caption{\textbf{Autonomous object navigation in changing environment}: The system is provided with a partial map of the environment containing an office room. A monitor is later placed in the office room and the system is tasked with finding it. In (a), the \textit{Object-Map-Search} baseline uses only a semantic object map without room labels. In (b), OrionNav uses the proposed hierarchical scene graph.}
    \label{fig:comparison}
    \vspace*{-0.5cm}
\end{figure}

\vspace{1.2mm}
\noindent \textbf{Frontier-Search.}
In our experiments, the \textit{Frontier-Search} baseline was largely ineffective, succeeding in only one of nine trials and requiring about three minutes. In contrast, OrionNav succeeded in all trials with substantially lower search times. This improvement stems from two key components: the hierarchical scene graph and the LLM-based planner. OrionNav continuously maintains a structured understanding of the environment through the scene graph, allowing the LLM planner to infer the most likely room containing the queried object and guide the robot to search there directly.
By contrast, \textit{Frontier-Search} lacks semantic reasoning and relies purely on frontier or random exploration, leading to inefficient searches and low success rates. Even when environmental changes initially prevent detection, OrionNav adapts by updating the scene graph and revising its plan, enabling successful object localization in subsequent attempts.

\vspace{1.2mm}
\noindent \textbf{Object-Map-Search.}
The key distinction between the \textit{Object-Map-Search} baseline and OrionNav is that the former relies on a semantic object map without room labels. This results in a success rate of about 56\%, compared to OrionNav’s 100\%, highlighting the benefit of the hierarchical scene graph. Fig.~\ref{fig:comparison} illustrates this difference in a scenario where both systems receive a partial map containing an office, a break room, and a long L-shaped corridor. The task is to locate a monitor that is absent during initial mapping and later returned to the office. In Fig.~\ref{fig:comparison}a, the baseline assumes the monitor is on a table and navigates to the nearest one in the break room, where no monitor is present. It then resorts to global exploration before eventually reaching the office and detecting the object. In contrast, OrionNav (Fig.~\ref{fig:comparison}b) uses the hierarchical scene graph to infer that a monitor is likely located in an office and navigates there directly, detecting it in a single step. As shown in Table~\ref{tab:comparison_experiment_duration}, even in successful trials, the baseline consistently required more time to complete the task than OrionNav.

\section{Conclusion}
\label{sec:conclusion}
In this paper, we introduced OrionNav, an open-architecture, end-to-end system for real-world autonomous navigation. The system continuously constructs an open-vocabulary hierarchical scene graph of the environment, which an LLM leverages to generate adaptive high-level plans, executed through a safety-critical low-level controller. Extensive real-world experiments on a Unitree Go2 quadruped demonstrate high success rates and practical feasibility. We also provide a detailed analysis of system performance, including module-level comparisons and failure cases. Overall, through this work, we take a step toward deploying robotic systems in real-world environments and aim to inspire further research in this direction.

\bibliographystyle{IEEEtran}
\bibliography{references}

@article{goswami2024floor,
  title={Floor Plan Based Active Global Localization and Navigation Aid for Persons With Blindness and Low Vision},
  author={Goswami, Raktim Gautam and Sinha, H and Amith, PV and Hari, J and Krishnamurthy, Prashanth and Rizzo, J and Khorrami, Farshad},
  journal={IEEE Robotics and Automation Letters},
  number={12},
  pages={11058--11065},
  year={2024},
  publisher={IEEE}
}

@inproceedings{nav2,
  title={The marathon 2: A navigation system},
  author={Macenski, Steve and Mart{\'\i}n, Francisco and White, Ruffin and Clavero, Jonatan Gin{\'e}s},
  booktitle={2020 IEEE/RSJ International Conference on Intelligent Robots and Systems (IROS)},
  pages={2718--2725},
  year={2020},
  organization={IEEE}
}

@article{llm_planning_survey,
  title={A Survey of Task Planning with Large Language Models},
  author={Zhai, Wenshuo and Liao, Jinzhi and Chen, Ziyang and Su, Bolun and Zhao, Xiang},
  journal={Intelligent Computing},
  volume={4},
  pages={0124},
  year={2025},
  publisher={AAAS}
}

@article{huang2024gamap,
  title={Gamap: Zero-shot object goal navigation with multi-scale geometric-affordance guidance},
  author={Huang, Hao and Hao, Yu and Wen, Congcong and Tzes, Anthony and Fang, Yi and others},
  journal={Advances in Neural Information Processing Systems},
  volume={37},
  pages={39386--39408},
  year={2024}
}

@inproceedings{wen2025zero,
  title={Zero-shot object navigation with vision-language models reasoning},
  author={Wen, Congcong and Huang, Yisiyuan and Huang, Hao and Huang, Yanjia and Yuan, Shuaihang and Hao, Yu and Lin, Hui and Liu, Yu-Shen and Fang, Yi},
  booktitle={International Conference on Pattern Recognition},
  pages={389--404},
  year={2025},
  organization={Springer}
}

@article{llama3,
  title={The llama 3 herd of models},
  author={Dubey, Abhimanyu and Jauhri, Abhinav and Pandey, Abhinav and Kadian, Abhishek and Al-Dahle, Ahmad and Letman, Aiesha and Mathur, Akhil and Schelten, Alan and Yang, Amy and Fan, Angela and others},
  journal={arXiv preprint arXiv:2407.21783},
  year={2024}
}

@article{RosinolVAHCSGC21,
  author       = {Antoni Rosinol and
                  Andrew Violette and
                  Marcus Abate and
                  Nathan Hughes and
                  Yun Chang and
                  Jingnan Shi and
                  Arjun Gupta and
                  Luca Carlone},
  title        = {Kimera: From {SLAM} to spatial perception with 3D dynamic scene graphs},
  journal      = {International Journal of Robotics Research},
  volume       = {40},
  number       = {12-14},
  pages        = {1510--1546},
  year         = {2021},
}

@article{slam_toolbox, 
    doi = {10.21105/joss.02783}, 
    url = {https://doi.org/10.21105/joss.02783}, 
    year = {2021}, 
    publisher = {The Open Journal}, 
    volume = {6}, 
    number = {61}, 
    pages = {2783}, 
    author = {Steve Macenski and Ivona Jambrecic}, 
    title = {{SLAM} Toolbox: {SLAM} for the dynamic world}, 
    journal = {Journal of Open Source Software} }

@inproceedings{convnext,
  title={A convnet for the 2020s},
  author={Liu, Zhuang and Mao, Hanzi and Wu, Chao-Yuan and Feichtenhofer, Christoph and Darrell, Trevor and Xie, Saining},
  booktitle={Proceedings of the Conference on Computer Vision and Pattern Recognition},
  pages={11966--11976},
   month = {June},
  year={2022},
  address = {New Orleans, LA, USA}
}

@article{dai2024sailing,
  title={Sailing Through Point Clouds: Safe Navigation Using Point Cloud Based Control Barrier Functions},
  author={Dai, Bolun and Khorrambakht, Rooholla and Krishnamurthy, Prashanth and Khorrami, Farshad},
  journal={{IEEE} Robotics and Automation Letters},
  year={2024},
  volume={9},
   number={9},
  pages={7731--7738},
}

@INPROCEEDINGS{10161317,
  author={Singh, Ishika and Blukis, Valts and Mousavian, Arsalan and Goyal, Ankit and Xu, Danfei and Tremblay, Jonathan and Fox, Dieter and Thomason, Jesse and Garg, Animesh},
  booktitle={Proceedings of the International Conference on Robotics and Automation}, 
  title={{ProgPrompt}: Generating Situated Robot Task Plans using Large Language Models}, 
  year={2023},
  month={May},
  pages={11523--11530},
  address={London, United Kingdom}
}

@InProceedings{huang2022inner,
  title = 	 {Inner Monologue: Embodied Reasoning through Planning with Language Models},
  author =       {Huang, Wenlong and Xia, Fei and Xiao, Ted and Chan, Harris and Liang, Jacky and Florence, Pete and Zeng, Andy and Tompson, Jonathan and Mordatch, Igor and Chebotar, Yevgen and others},
  booktitle = 	 {Proceedings of the Conference on Robot Learning},
  pages = 	 {1769--1782},
  year = 	 {2023},
  month = 	 {December},
  address= {Atlanta, GA, USA}
}

@inproceedings{RadfordKHRGASAM21,
  author       = {Alec Radford and
                  Jong Wook Kim and
                  Chris Hallacy and
                  Aditya Ramesh and
                  Gabriel Goh and
                  Sandhini Agarwal and
                  Girish Sastry and
                  Amanda Askell and
                  Pamela Mishkin and
                  Jack Clark and
                  Gretchen Krueger and
                  Ilya Sutskever},
  title        = {Learning Transferable Visual Models From Natural Language Supervision},
  booktitle    = {Proceedings of the International Conference on Machine Learning},
  month        = {July},
  address      = {Vienna, Austria},
  volume       = {139},
  pages        = {8748--8763},
year= {2021}
}

@inproceedings{Ahn2022DoAI,
  title={Do As {I} Can, Not As {I} Say: Grounding Language in Robotic Affordances},
  author={Michael Ahn and Anthony Brohan and Noah Brown and Yevgen Chebotar and Omar Cortes and Byron David and Chelsea Finn and Keerthana Gopalakrishnan and Karol Hausman and Alexander Herzog and others},
  booktitle={Proceedings of the Conference on Robot Learning},
  year={2022},
  month = {December},
  address= {Auckland, New Zealand},
  pages = {287--318}
}

@inproceedings{rana2023sayplan,
  title={Sayplan: Grounding large language models using 3d scene graphs for scalable robot task planning},
  author={Rana, Krishan and Haviland, Jesse and Garg, Sourav and Abou-Chakra, Jad and Reid, Ian and Suenderhauf, Niko},
  booktitle={Proc. Conference on Robot Learning},
  year={2023},
month={November},
address={Atlanta, GA, USA},
pages={23--72}
}

@inproceedings{rajvanshi2024saynav,
  title={Saynav: Grounding large language models for dynamic planning to navigation in new environments},
  author={Rajvanshi, Abhinav and Sikka, Karan and Lin, Xiao and Lee, Bhoram and Chiu, Han-Pang and Velasquez, Alvaro},
  booktitle={Proceedings of the International Conference on Automated Planning and Scheduling},
  pages={464--474},
  year={2024},
  month={June},
  address={Banff, AB, Canada}
}

@ARTICLE{10632580,
  author={Honerkamp, Daniel and Büchner, Martin and Despinoy, Fabien and Welschehold, Tim and Valada, Abhinav},
  journal={{IEEE} Robotics and Automation Letters}, 
  title={Language-Grounded Dynamic Scene Graphs for Interactive Object Search With Mobile Manipulation}, 
  year={2024},
  volume={9},
  number={10},
  pages={8298--8305}}

@article{liu2024okrobot,
  title={{OK-Robot}: What Really Matters in Integrating Open-Knowledge Models for Robotics},
  author={Liu, Peiqi and Orru, Yaswanth and Paxton, Chris and Shafiullah, Nur Muhammad Mahi and Pinto, Lerrel},
  journal={arXiv preprint arXiv:2401.12202},
  year={2024}
}

@inproceedings{cai2024bridging,
  title={Bridging zero-shot object navigation and foundation models through pixel-guided navigation skill},
  author={Cai, Wenzhe and Huang, Siyuan and Cheng, Guangran and Long, Yuxing and Gao, Peng and Sun, Changyin and Dong, Hao},
  booktitle={Proceedings of the International Conference on Robotics and Automation},
  pages={5228--5234},
  year={2024},
  month={May},
  address={Yokohama, Japan}
}

@inproceedings{werby2024hierarchical,
  AUTHOR    = {Abdelrhman Werby AND Chenguang Huang AND Martin Büchner AND Abhinav Valada AND Wolfram Burgard}, 
    TITLE     = {{Hierarchical Open-Vocabulary 3D Scene Graphs for Language-Grounded Robot Navigation}}, 
    BOOKTITLE = {Proceedings of the Robotics: Science and Systems}, 
    YEAR      = {2024}, 
    ADDRESS   = {Delft, Netherlands}, 
    MONTH     = {July}, 
}

@inproceedings{gu2024conceptgraphs,
  title={Conceptgraphs: Open-vocabulary 3d scene graphs for perception and planning},
  author={Gu, Qiao and Kuwajerwala, Ali and Morin, Sacha and Jatavallabhula, Krishna Murthy and Sen, Bipasha and Agarwal, Aditya and Rivera, Corban and Paul, William and Ellis, Kirsty and Chellappa, Rama and others},
  booktitle={Proceedings of the International Conference on Robotics and Automation},
  pages={5021--5028},
  year={2024},
  month={May},
  address={Yokohama, Japan}
}

@article{achiam2023gpt,
  title={Gpt-4 technical report},
  author={Achiam, Josh and Adler, Steven and Agarwal, Sandhini and Ahmad, Lama and Akkaya, Ilge and Aleman, Florencia Leoni and Almeida, Diogo and Altenschmidt, Janko and Altman, Sam and Anadkat, Shyamal and others},
  journal={arXiv preprint arXiv:2303.08774},
  year={2023}
}

@ARTICLE{Maggio2024Clio,
    title={Clio: Real-time Task-Driven Open-Set 3D Scene Graphs},
    author={Maggio, Dominic and Chang, Yun and Hughes, Nathan and Trang, Matthew and
    Griffith, Dan and Dougherty, Carlyn and Cristofalo, Eric and
    Schmid, Lukas and Carlone, Luca},
    journal={{IEEE} Robotics and Automation Letters},
    year={2024},
    volume={9},
    number={10},
    pages={8921-8928}
}

@article{ZhangHXZ22,
  author       = {Tianyao Zhang and
                  Xiaoguang Hu and
                  Jin Xiao and
                  Guofeng Zhang},
  title        = {A survey of visual navigation: From geometry to embodied {AI}},
  journal      = {Engineering Applications of Artificial Intelligence},
  volume       = {114},
  pages        = {105036},
  year         = {2022},
}

@article{duan2022survey,
  title={A survey of embodied ai: From simulators to research tasks},
  author={Duan, Jiafei and Yu, Samson and Tan, Hui Li and Zhu, Hongyuan and Tan, Cheston},
  journal={{IEEE} Transactions on Emerging Topics in Computational Intelligence},
  volume={6},
  number={2},
  pages={230--244},
  year={2022}
}

@inproceedings{SchultEHLTL23,
  author       = {Jonas Schult and
                  Francis Engelmann and
                  Alexander Hermans and
                  Or Litany and
                  Siyu Tang and
                  Bastian Leibe},
  title        = {Mask3D: Mask Transformer for 3D Semantic Instance Segmentation},
  booktitle={Proceedings of the International Conference on Robotics and Automation},
  pages        = {8216--8223},
  year         = {2023},
  month        = {May},
  address      = {London, UK}
}

@inproceedings{fcclip,
  author       = {Qihang Yu and
                  Ju He and
                  Xueqing Deng and
                  Xiaohui Shen and
                  Liang{-}Chieh Chen},
  title        = {Convolutions Die Hard: Open-Vocabulary Segmentation with Single Frozen
                  Convolutional {CLIP}},
  booktitle={Proceedings of the Advances in Neural Information Processing Systems},
  month={December},
  year={2023},
  address={New Orleans, LA, USA},
}

@inproceedings{YuanLZLC24,
  author       = {Haobo Yuan and
                  Xiangtai Li and
                  Chong Zhou and
                  Yining Li and
                  Kai Chen and
                  Chen Change Loy},
  title        = {Open-Vocabulary {SAM:} Segment and Recognize Twenty-thousand Classes
                  Interactively},
  booktitle={Proceedings of the European Conference on Computer Vision},
  month={September},
  year={2024},
  address= {Milan, Italy},
  pages={}
}

@inproceedings{DingYXZBQ23,
  author       = {Runyu Ding and
                  Jihan Yang and
                  Chuhui Xue and
                  Wenqing Zhang and
                  Song Bai and
                  Xiaojuan Qi},
  title        = {{PLA:} Language-Driven Open-Vocabulary 3D Scene Understanding},
  booktitle    = {Proc. International Conference on Computer Vision and Pattern Recognition},
  address      = {Vancouver, BC, Canada},
  month        = {June},
  pages        = {7010--7019},
  year         = {2023},
}

@article{WangL24f,
  author       = {Hao Wang and
                  Minghui Li},
  title        = {A New Era of Indoor Scene Reconstruction: {A} Survey},
  journal      = {{IEEE} Access},
  volume       = {12},
  pages        = {110160--110192},
  year         = {2024},
}

@article{YinXLCXSSW24,
  author       = {Huan Yin and
                  Xuecheng Xu and
                  Sha Lu and
                  Xieyuanli Chen and
                  Rong Xiong and
                  Shaojie Shen and
                  Cyrill Stachniss and
                  Yue Wang},
  title        = {A Survey on Global LiDAR Localization: Challenges, Advances and Open
                  Problems},
  journal      = {International Journal of Computer Vision},
  volume       = {132},
  number       = {8},
  pages        = {3139--3171},
  year         = {2024},
}

@article{WangTCXD24,
  author       = {Yanan Wang and
                  Yaobin Tian and
                  Jiawei Chen and
                  Kun Xu and
                  Xilun Ding},
  title        = {A Survey of Visual {SLAM} in Dynamic Environment: The Evolution From
                  Geometric to Semantic Approaches},
  journal      = {{IEEE} Transactions on Instrumentation and Measurement},
  volume       = {73},
  pages        = {1--21},
  year         = {2024},
}

@article{WuLXYDY0ZT0GT24,
  author       = {Jianzong Wu and
                  Xiangtai Li and
                  Shilin Xu and
                  Haobo Yuan and
                  Henghui Ding and
                  Yibo Yang and
                  Xia Li and
                  Jiangning Zhang and
                  Yunhai Tong and
                  Xudong Jiang and
                  Bernard Ghanem and
                  Dacheng Tao},
  title        = {Towards Open Vocabulary Learning: {A} Survey},
  journal      = {{IEEE} Transactions on Pattern Analysis and Machine Intelligence},
  volume       = {46},
  number       = {7},
  pages        = {5092--5113},
  year         = {2024}
}

@inproceedings{yokoyama2024vlfm,
  title={Vlfm: Vision-language frontier maps for zero-shot semantic navigation},
  author={Yokoyama, Naoki and Ha, Sehoon and Batra, Dhruv and Wang, Jiuguang and Bucher, Bernadette},
  booktitle={Proceedings of the International Conference on Robotics and Automation},
  pages={42--48},
  address={Yokohama, Japan},
  month={May},
  year={2024},
  organization={IEEE}
}

@inproceedings{zhou2023esc,
  title={{ESC}: Exploration with soft commonsense constraints for zero-shot object navigation},
  author={Zhou, Kaiwen and Zheng, Kaizhi and Pryor, Connor and Shen, Yilin and Jin, Hongxia and Getoor, Lise and Wang, Xin Eric},
  booktitle={International Conference on Machine Learning},
  pages={42829--42842},
  year={2023},
  month={July},
 address = {Honolulu, HI}
}

@INPROCEEDINGS{Chen-RSS-23, 
    AUTHOR    = {Junting Chen AND Guohao Li AND Suryansh Kumar AND Bernard Ghanem AND Fisher Yu}, 
    TITLE     = {{How To Not Train Your Dragon: Training-free Embodied Object Goal Navigation with Semantic Frontiers}}, 
    BOOKTITLE = {Proceedings of the Robotics: Science and Systems}, 
    YEAR      = {2023}, 
    ADDRESS   = {Daegu, Republic of Korea}, 
    MONTH     = {July}
}

@INPROCEEDINGS{11128486,
  author={Devarakonda, Venkata Naren and Kaypak, Ali Umut and Yuan, Shuaihang and Krishnamurthy, Prashanth and Fang, Yi and Khorrami, Farshad},
  booktitle={Proc. International Conference on Robotics and Automation}, 
  title={{MultiTalk}: Introspective and Extrospective Dialogue for Human-Environment-LLM Alignment}, 
  year={2025},
  month={May},
  volume={},
  number={},
  pages={10737-10743},
  address={Atlanta, GA, USA}}

\end{document}